\documentclass{article}

    \PassOptionsToPackage{numbers, compress}{natbib}

    \usepackage[preprint]{neurips_2019}

\usepackage[utf8]{inputenc} 
\usepackage[T1]{fontenc}    
\usepackage{hyperref}       
\usepackage{url}            
\usepackage{booktabs}       
\usepackage{amsfonts}       
\usepackage{nicefrac}       
\usepackage{microtype}      

\usepackage{amsmath}
\usepackage{xcolor}

\usepackage{footnote}
\makesavenoteenv{tabular}
\makesavenoteenv{table}

\DeclareMathOperator*{\argmax}{arg\,max}

\title{Purifying Adversarial Perturbation with Adversarially Trained Auto-encoders}

\author{%
  Hebi Li \\
  Iowa State University\\
  \texttt{hebi@iastate.edu} \\
   \And
   Qi Xiao \\
   Iowa State University\\
   \texttt{qxiao@iastate.edu} \\
   \AND
   Shixin Tian \\
   Iowa State University\\
   \texttt{stian@iastate.edu} \\
   \And
   Jin Tian \\
   Iowa State University\\
   \texttt{jtian@iastate.edu} \\
}

\begin{document}

\maketitle

\begin{abstract}
  Machine learning models are vulnerable to adversarial
  examples. Iterative adversarial training has shown promising results
  against strong white-box attacks. However, adversarial training is
  very expensive, and every time a model needs to be protected, such
  expensive training scheme needs to be performed. In this paper, we
  propose to apply iterative adversarial training scheme to an
  external auto-encoder, which once trained can be used to protect
  other models directly. We empirically show that our model
  outperforms other purifying-based methods against white-box attacks,
  and transfers well to directly protect other base models with
  different architectures.

\end{abstract}

\section{Introduction}
Deep neural models have shown tremendous success in recent years in
many machine learning tasks. However, like other machine learning
models, researchers have found neural models are vulnerable to
adversarial attacks, where small perturbation of input flips the
model's output with high confidence.

Many defense techniques have been proposed, but have achieved limited
success on white-box attacks. Early approaches such
as~\cite{2016-SP-Papernot-Distillation, 2017-SIGSAC-Meng-Magnet,
  2018-CVPR-Liao-Defense} only defend against black-box attacks and/or
oblivious attacks~\cite{2017-AISec-Carlini-Adversarial}, where
attackers are not aware of the presence of defense. On the contrary,
white-box attackers have full knowledge of both the base model and the
defense. To defend against such strong white-box attacks, researchers
tried to hide the input gradients that are used by first-order
attackers. Such techniques include making some operations in the model
non-differentiable~\cite{2018-ICLR-Buckman-Thermometer,
  2018-ICLR-Guo-Countering, 2018-ICLR-Ma-Characterizing}, adding
stochastic to the gradients~\cite{2018-ICLR-Dhillon-Stochastic,
  2018-ICLR-Xie-Mitigating}, and exploring gradients
vanishing/exploding~\cite{2018-ICLR-Song-Pixeldefend}. However, these
techniques can be side-stepped by approximating the true
gradients~\cite{2018-ICML-Athalye-Obfuscated}, and as a result
provided little robustness to white-box attacks.

The most successful defense against white-box attacks is iterative
adversarial
training~\cite{2018-ICLR-Madry-Towards,2017-ICLR-Kurakin-Adversarial}.
Adversarial training~\cite{2015-ICLR-Goodfellow-Explaining} is a
defense technique that augments training data with adversarially
generated examples. Madry et al.~\cite{2018-ICLR-Madry-Towards} showed
that when applying adversarial training in an iterative manner, the
model can achieve state-of-the-art defense against white-box attacks.

However, iterative adversarial training introduces significant
computational overhead, as it generates adversarial examples
on-the-fly in every training iteration. Depending on the attack
algorithm used, the overhead is typically 10-30x, and it is hard for
the technique to scale to
ImageNet~\cite{2017-ICLR-Kurakin-Adversarial,2018-ICLR-Tramer-Ensemble}.
In addition, iterative adversarial training achieves robustness
through modifying the parameters of the base classifier under
protection. In other words, the robustness is embedded inside the
model, and cannot be easily transferred to other models. Every time a
model needs to be protected, such expensive training scheme needs to
be performed.

On the other hand, some approaches do not modify the base model, but
purify the input. Representative approaches include
DefenseGAN~\cite{2018-ICLR-Samangouei-Defense},
MagNet~\cite{2017-SIGSAC-Meng-Magnet} and several
others~\cite{2019-ICASSP-Jin-APE,2019-Preprint-Hwang-PuVAE,2018-ICLR-Song-Pixeldefend,2018-CVPR-Liao-Defense}.
We call such approaches the \emph{purifying-based methods}. Since the
data domain does not change after purification, such models, once
trained, can be directly used to protect different base models.  We
refer such property as ``transferability'' or ``generalizability'' in
this paper, and use the two terms interchangeably.
However, current purifying-based approaches have shown limited success
in white-box attack.  For example,
MagNet~\cite{2017-SIGSAC-Meng-Magnet} and
HGD~\cite{2018-CVPR-Liao-Defense} has been shown to provide no
white-box
security~\cite{2017-Preprint-Carlini-Magnet,2018-Preprint-Athalye-Robustness}. And
it has been shown PixelDefend~\cite{2018-ICLR-Song-Pixeldefend} and
DefenseGAN~\cite{2018-ICLR-Samangouei-Defense} could be bypassed by
simple gradient approximation of the vanished gradients
\cite{2018-ICML-Athalye-Obfuscated}.


In this paper, we study the possibility to combine the robustness of
iterative adversarial training with the generalizability of
purifying-based approaches. In particular, we propose to apply
iterative adversarial training scheme to train only an external
auto-encoder instead of the underlying base model. The auto-encoder is
trained to perform two functionalities at the same time. On one hand,
it should not modify clean input so that clean images can still be
classified correctly by the model; on the other hand, the auto-encoder
is trained with adversarial examples in iterative manner, such that it
removes the adversarial perturbation before feeding inputs to the base
classifier, even against strong white-box attacks.

We empirically evaluate the robustness of our model against three
threat models on MNIST and CIFAR10 dataset, and show state-of-the-art
white-box robustness compared with other purifying-based methods. We
also demonstrate the transferability of our approach by showing that,
the auto-encoder trained on one model performs well when plugged in
other models with different architectures. To further improve the
transferability of the model, we propose to apply ensemble training
across multiple base models, and show empirically improved
performance.

In summary, we make the following contributions:

\begin{itemize}
\item We propose a novel purifying-based method to protect neural
  models against adversarial examples. In particular, we apply
  iterative adversarial training to external auto-encoders. We
  empirically show that our method outperforms other purifying-based
  method on strong white-box attacks.
\item We show that our trained auto-encoder on one base model can be
  used directly to protect other base models with high accuracy,
  achieving model transferabiltiy. We further propose to apply
  ensemble training across different base models, and empirically show
  the performance boost on model transferability.
\end{itemize}

\section{Background}

\subsection{Model and notations}
We assume the pre-trained base model is an $m$-class classifier in the
form of neural networks.  Following the notation used in the
literature~\cite{2017-SP-Carlini-Towards,2018-ICLR-Madry-Towards}, let
$F_\theta(x)=y$ be an $m$-class neural network classifier that accepts
an $n$-dimensional image input $x \in \mathbb{R}^n$, and produces
$y \in [0,1]^m$ as the probability of prediction.  The output $y$ is
computed using the softmax function to ensure $\sum y_i = 1$.  The
input $x$ is classified as $C(x)=\argmax_i(y_i)$.  The model parameter
$\theta$ is trained to minimize some loss $L(\theta; x, y)$, typically
the cross-entropy loss, i.e.  $L(\theta; x, y) = H(F_\theta(x), y)$.

\subsection{Adversarial attacks}
An adversarial example of $x$ for the model $F(x)$ is $x'$ that is
similar to $x$ in some distance measure $D(x,x')$, such that
$C(x') \ne C(x)$. The difference $\delta = x' - x$ is called
adversarial perturbation.

\paragraph{Fast Gradient Sign Method (FGSM)} Given an image $x$ and
its corresponding true label $y$, the FGSM
attack~\cite{2015-ICLR-Goodfellow-Explaining} computes the adversarial
example as:

\begin{equation}
x' = x + \epsilon \cdot sign(\nabla_x L(\theta; x,y))
\end{equation}

where $\epsilon$ is a small constant denoting the amount of
perturbation.  FGSM is designed to be very fast to generate
proof-of-concept adversarial examples.

\paragraph{Projected Gradient Descent (PGD)} Also known as Basic
Iterative Method (BIM) or Iterative FGSM
($FGSM^k$)~\cite{2017-ICLR-Kurakin-Adversarial,
  2018-ICLR-Madry-Towards}, PGD is a more powerful, multi-step variant
of FGSM. In each iteration, the input is projected to be in the valid
range [0,1] and within allowed perturbation range $S$. Formally,

\begin{equation}
x^{t+1} = Proj_{x+S} (x_t + \epsilon \cdot sign(\nabla_{x^t} L(\theta; x^t,y)))
\end{equation}

The Randomized Fast Gradient Sign Method
(RAND+FGSM)~\cite{2018-ICLR-Tramer-Ensemble} is the one iteration
special case of PGD, and thus is strictly weaker than PGD. In
particular, PGD has been suggested to be the universal first-order
$\ell_\infty$-bounded attack~\cite{2018-ICLR-Madry-Towards}.



\paragraph{Carlini-Wagner (CW) $\ell_2$ attack} CW $\ell_2$
attack~\cite{2017-SP-Carlini-Towards} is an optimization based
technique as opposed to the above gradient sign methods. The
optimization problem is formulated as:


\begin{equation}
  \begin{aligned}
    & \mathop{\text{min}}_\delta & & \|\delta\|_2 + c \cdot f(x+\delta)\\
    & \text{s.t.} & & x+\delta \in [0,1]^n
  \end{aligned}
\end{equation}

where $f$ is a function specially designed such that $f(x')<0$ if and
only if $C(x')=t$, with $t$ being the target adversarial class
label. This optimization approach is shown to be much stronger than
gradient-based ones, but with the price of computational overhead due
to the optimization steps.





\subsection{Iterative adversarial training}

The idea of iterative adversarial
training~\cite{2018-ICLR-Madry-Towards,2017-ICLR-Kurakin-Adversarial}
is to train the classifier with adversarial examples generated by some
attacks during the training process.  Formally, iterative adversarial
training is formulated as a min-max optimization problem:

\begin{equation} \label{eq:advtrain}
\mathop{\text{min}}_{\theta} \mathop{\mathbb{E}}_{(x,y) \in \chi} [~\mathop{\text{max}}_{\delta \in S} L(\theta; x+\delta, y ~)]
\end{equation}

where $\chi$ is the data distribution, $S \in \mathbb{R}^n$ is some
allowed perturbations, typically set to $\ell_\infty$-ball in the
literature of adversarial training due to runtime efficiency of
performing FGSM and PGD.


\subsection{Threat Model}


Following terminology in the
literature~\cite{2017-CCS-Papernot-Practical,2017-AISec-Carlini-Adversarial},
we consider the following threat models based on how much information
the attacker has access to.

\begin{itemize}
\item \textbf{black-box attack}: the attacker knows neither the model
  nor the defense~\cite{2017-CCS-Papernot-Practical}. The attacker is
  allowed to query the model (with defense in place) for the outputs
  of given inputs, collect data, and train a substitute model, and
  perform attack on the substitute model to generate adversarial
  examples. These adversarial examples are then tested on the
  black-box model.
\item \textbf{oblivious attack}: the attacker has access to the full base 
  classification model, including architecture and parameters. But the
  attacker does \emph{not} know the presence of the defense, thus
  adversarial examples are only generated for the base model.
\item \textbf{white-box attack}: the attacker has full access to the
  base model as well as the defense, including all architectures and
  parameters. The adversarial examples are generated against the
  model with defense in place.
\end{itemize}

We remark that in the literature there has been inconsistent usage of
oblivious and white-box threat models. Many papers report only
oblivious attack results while claiming white-box ones. As suggested
by the literature, white-box attacks are much stronger and more
realistic threat model than oblivious
attacks~\cite{2017-AISec-Carlini-Adversarial}.


\section{Approach}

Given a classifier model $F_\theta(x)$, we add an auto-encoder
$AE_\phi(x)$ to preprocess the input before classifying it. Thus the
full model is

\begin{equation}
M_{\theta, \phi}(x) = \text{F}_\theta(AE_\phi(x))
\end{equation}

Let $x$ denote the clean image, $x+\delta$ denote the adversarial
image with perturbation $\delta$.  Instead of minimizing adversarial
loss over the classifier model parameter $\theta$ as was done in
Eq.~\ref{eq:advtrain}, in this paper, we assume that the classifier
model $F_\theta(x)$ has been pre-trained with clean images and
$\theta$ is fixed, and we train only the auto-encoder parameter $\phi$
with the following adversarial objective:

\begin{equation} \label{eq:A2}
  \mathop{\text{min}}_{\phi} \mathop{\mathbb{E}}_{(x,y) \in \chi}
  \Big[~\mathop{\text{max}}_{\delta \in S} L\Big(\theta; AE_\phi(x+\delta), y \Big)\Big]
\end{equation}

It turns out training only on this adversarial objective may lead to
decrease of clean image accuracy. Thus, we propose to combine the
adversarial objective with the reconstruction error of clean image and
train the auto-encoder parameter $\phi$ with the following objective:

\begin{equation} \label{eq:C0}
  \mathop{\text{min}}_{\phi} \mathop{\mathbb{E}}_{(x,y) \in \chi}
  \Big[
    ~\mathop{\text{max}}_{\delta \in S} L\Big(\theta; AE_\phi(x+\delta), y \Big) + \lambda D\Big(AE_\phi(x), x \Big)
    \Big]
\end{equation}

where $D$ is some distance metric, in particular cross-entropy is used
in our experiment. $\lambda \ge 0$ is a hyper parameter to trade-off
the two error terms.

\subsection{Auto-encoder architecture}
\label{sec:autoencoder}

Auto-encoder is originally proposed as
denoiser~\cite{2008-ICML-Vincent-Extracting,2010-JMLR-Vincent-Stacked}. It
consists of an encoder and a decoder. Typical auto-encoders have a
bottleneck layer in the middle that is much lower dimensional than the
input, so that only useful information is retained after training. The
training objective is the reconstruction error between encoder's input
and decoder's output under Gaussian noise.  When applying in image
domain, encoder and decoder are typically stacked convolutional
layers.

For more challenging datasets, the bottleneck layer causes the
denoised image to be blur, and as a consequence the model experiences
a significant drop in classification accuracy on clean images. This
creates an upper limit on the defense performance against adversarial
inputs. Inspired by previous work by Liao et
al.~\cite{2018-CVPR-Liao-Defense}, we use
U-net~\cite{2015-Unknown-Ronneberger-Unet} as the auto-encoder
architecture to solve this problem. U-net was originally proposed for
image segmentation tasks. It adds lateral connections from encoder
layers to their corresponding decoder layers. This way, the fine-level
details of the image are retained, which helps dramatically with the
fidelity of the auto-encoder.


\subsection{Ensemble training across models}



The optimization in Eq.~\ref{eq:C0} has the base model in place.  A
natural idea to increase its generalizability is to train the model on
multiple base models. Based on this insight, we propose to apply
ensemble training to the auto-encoder across different base models.
Such ensemble training acts as a regularizer, so that the auto-encoder
generalizes across models.  Formally, given a list of base models
$F_1, \ldots, F_k$ with loss term $L_1, \ldots, L_k$, we optimize the
following problem:

\begin{equation} \label{eq:ensemble}
  \mathop{\text{min}}_{\phi}
  \sum_{i=1}^k
  \mathop{\mathbb{E}}_{(x,y) \in \chi}
  \Big[~\mathop{\text{max}}_{\delta \in S} L_i\Big(\theta_i; AE_\phi(x+\delta), y \Big)
  + \lambda D\Big(AE_\phi(x), x \Big)
  \Big]
\end{equation}


\section{Experiment}
We implement the model using Tensorflow~\cite{abadi2016tensorflow} and
Cleverhans~\cite{papernot2018cleverhans} framework. Our source code
will be freely available, and is attached in the supplementary
material for a preview. The experiment is performed on a single RTX
2080Ti graphic card.
\subsection{Experiment setup}
\paragraph{Dataset and Model architecture}

We apply our approach to MNIST \cite{lecun1998mnist} and CIFAR10
\cite{krizhevsky2009learning} dataset. MNIST consists of a training
set of 60,000 and testing set of 10,000 28x28x1 images of handwritten
digits with 10 classes. CIFAR10 contains a training set of 50,000 and
testing set of 10,000 32x32x3 RGB images of objects with 10
classes. The pixel value is normalized into [0,1] by dividing
255. During training, 10\% of training data is hold out as validation
data. All tests are evaluated on testing data.

For MNIST, we use a simple CNN as base model with four convolutional
layers, containing 32,32,64,64 channels, all with 3x3 filters. The
architecture detail is depicted as model X in Table \ref{tab:arch}.
This model achieves 99.6\% accuracy on clean data.  We use a simple
auto-encoder model where the encoder consists of two convolutional
blocks, each with a 32 channel, 3x3 convolutional layer, a batch
normalization layer, a ReLU activation layer, and a 2x2 max pooling
layer.  The decoder consists of two convolutional blocks as well, each
with a 32-channel 3x3 convolutional layer, a ReLU activation layer,
and a 2x2 up sampling layer.  The output goes through one last
convolutional layer of 3 channels activated by sigmoid function to
ensure that the decoded image is in valid range [0,1], and the shape
is the same as input.

For CIFAR10, we use a standard ResNetV2 with 29 convolutional layers.
We apply standard data augmentation, including random crops and flips
as were done in~\cite{2018-ICLR-Madry-Towards}. This network achieves
89\% on clean test images.  The
U-net~\cite{2015-Unknown-Ronneberger-Unet} described in
Sec~\ref{sec:autoencoder} is used as the auto-encoder
architecture. Specifically, the encoder contains 5 convolutional
layers, with channel size 32,64,128,256,512 and 3x3 kernel
filters. The decoder consists of 5 convolutional layers with channel
size 256,128,64,32,3 and 3x3 kernels. There are residual connections
to stack the layer in the encoder onto the corresponding layer in the
decoder.
The final decoded image is produced by a sigmoid activation function
to produce valid image in range [0,1].

\paragraph{Training Details}

We train the auto-encoder using the combined adversarial loss and
clean image reconstruction loss in Eq.~\ref{eq:C0}. For MNIST,
$\lambda$ is set to 0, because the reconstruction term is not
necessary for the model to reach high accuracy on clean image. For
CIFAR10, $\lambda$ is set to 1.  During iterative adversarial
training, we use PGD to generate adversarial examples. For MNIST
model, we use 40 iterations of PGD with step size 0.01 and a maximum
distortion of 0.3. For CIFAR10, we uses 10 iterations of PGD with step
size \nicefrac{2}{255}~maximum distortion \nicefrac{8}{255}.

All models are trained using Adam optimizer with learning rate of 1e-3
and learning rate decay of a factor of 0.5 when the validation loss
does not improve over 4 epochs. Training early stops if the validation
loss does not improve over 10 epochs, and the best model is
restored. It typically takes 30 to 50 epochs to train a model. We
observe consistent performance during development, and report results
for a single run.

\paragraph{Threat model}
We evaluate attacks in all three kinds of threat models. As a recap,
the white-box attack is applied to the entire model (CNN with
auto-encoder in place), and the oblivious attack is applied to CNN
model only, and the generated adversarial examples are used to attack
the model with defense. For black-box attack, we
follows~\cite{2017-CCS-Papernot-Practical} and train a substitute
model using the 150 holdout data samples in testing dataset as a seed,
and query our entire model (CNN with Auto-encoder) as the black-box
model for 6 epochs. Then, we apply attack on the substitute model in a
white-box manner, and test the resulting adversarial examples on the
black-box model. We test different models as the substitute model, and
do not find statistically significant difference, thus we report one
result where the substitute model has the same architecture as the
black-box model.
\paragraph{Applied attacks}
We evaluate both $\ell_\infty$-bounded attacks (FGSM and PGD) and
$\ell_2$-bounded attacks (CW $\ell_2$).  For FGSM, we set the
distortion to be $\epsilon=0.3$ for MNIST and
$\epsilon=\nicefrac{8}{255}$ for CIFAR10. For PGD, we use the same
attack parameters as those used during training.  We do not present
the $CW \ell_\infty$ attack, as the PGD has been shown to be a
universal first-order $\ell_\infty$-bounded attack.

For $\ell_2$-bounded attack, we use the untargeted CW $\ell_2$ attack
with learning rate 0.2 and max iteration 1000.  We remark that
previously, adversarial training is only evaluated on
$\ell_\infty$-bounded attacks~\cite{2018-ICLR-Madry-Towards}. That's
because adversarial training used $\ell_\infty$ attacks during
training, and thus only claims $\ell_\infty$ security.  In this
evaluation, we show that models trained with $\ell_\infty$ attacks can
also provide substantial robustness against CW $\ell_2$ attack, though
only for simple MNIST dataset.

For each attack, we perform adversarial perturbation on 1000 random
samples from testing dataset, and report the model accuracy on these
adversarial examples.


\subsection{Models in comparison}

We compare our model (denoted as AdvAE) with three representative
models, HGD, DefenseGAN, and iterative adversarial training (denoted
It-Adv-train). We compare with
DefenseGAN~\cite{2018-ICLR-Samangouei-Defense} because it is the
best-performing purifying-based technique against white-box attacks in
practice.  We choose to compare with HGD~\cite{2018-CVPR-Liao-Defense}
for two reasons. First, it is closely related to our approach, thus
the comparison can clear show the advantage of applying iterative
training scheme. Second, it acts as a representative of approaches
which claim only oblivious attack security but not white-box one, and
we show they provide little white-box robustness. It-Adv-train is not
a purifying-based method, but since our technique also builds atop
this technique, we compare the performance to see if applying
adversarial training on part of the model can achieve similar
robustness.  We omit other earlier approaches such as
MagNet~\cite{2017-SIGSAC-Meng-Magnet} and
PixelDefend~\cite{2018-ICLR-Song-Pixeldefend}, because they have been
shown to perform worse~\cite{2018-ICLR-Samangouei-Defense}, and does
not provide white-box security~\cite{2018-ICML-Athalye-Obfuscated,
  2017-Preprint-Carlini-Magnet}.



When performing white-box attacks on DefenseGAN, we follows the BPDA
algorithm suggested by Athalye et
al.~\cite{2018-ICML-Athalye-Obfuscated} to approximate the vanished
gradients. In particular, we directly set the gradient of the
``projection step'' of DefenseGAN to 1 during back-propagation when
calculating the input gradient.  We train GAN for 200,000
iterations. The DefenseGAN projection step is performed with gradient
descent iteration L=200 and random restart R=10, as was evaluated in
the original paper.





  

\subsection{MNIST results}
\begin{table}
\small
  \caption{MNIST accuracy results on three threat models.}
  \label{tab:mnist}
  \centering
  \begin{tabular}{lr|rr|rrr|r}
    \toprule
    & & oblivious & black-box & \multicolumn{4}{c}{white-box (left: purifying-based)}                 \\
    \cmidrule(r){3-3} \cmidrule(r){4-4} \cmidrule(r){5-8}
    
    Attacks & No def & AdvAE & AdvAE & \textbf{AdvAE} & HGD & DefGAN\footnote{This result is different from the performance reported in original DefenseGAN paper, because we apply BPDA attack~\cite{2018-ICML-Athalye-Obfuscated} to approximate the vanished gradient.}  & It-Adv-train \\
    
    \midrule
No attack & 99.6 &  &  & 99.1 & 98.3 & 92.2 & 99.5\\
FGSM $\ell_\infty$ & 41.1 & 96.5 & 96.8 & \textbf{96.9} & 57.1 & 87.8 & \textbf{98.0}\\
PGD $\ell_\infty$ & 1.6 & 97.1 & 97.6 & \textbf{94.8} & 1.3 & 44.0 & \textbf{97.2}\\
CW $\ell_2$ & 0.0 & 96.6 & 97.3 & \textbf{82.7} & 0.0 & 36.5 & \textbf{93.0}\\
    
    \bottomrule
  \end{tabular}
\end{table}

The MNIST experiment results against three threat models are shown in
Table \ref{tab:mnist}. The CNN model achieves 99.6\% accuracy on clean
data. As a baseline, when no defense is applied, the model is very
vulnerable against all attacks. By comparison, the oblivious results
of AdvAE show that adding our auto-encoder can indeed protect the
model very well without modifying the model itself. The black-box
accuracy does not drop, showing that our model does not suffer from
adversarial examples transferred from substitute models.

For white-box results, compared to HGD which also features an external
denoiser, we show that, by applying iterative adversarial training
scheme instead of a single adversarial training step, we can
tremendously boost the white-box security, from 1.3\% accuracy to
94.8\% for PGD attack and from 0\% to 82.7\% for CW $\ell_2$
attack. We show that DefenseGAN performs poorly under BPDA attack,
with 44.0\% and 36.5\% accuracy against PGD and CW $\ell_2$,
respectively. Thus, we claim ours as the new state-of-the-art
purifying-based model against white-box attacks. Comparing with
It-Adv-train, AdvAE performs worse but still comparable. This is as
expected, because AdvAE restricts itself by not modifying the base
model. This result shows that AdvAE successfully introduces the
white-box robustness from It-Adv-train into purifying-based
approaches.


\subsection{MNIST transferability results}

\begin{table}
\small
  \caption{Alternative model architectures. Each convolutional layer is activated by ReLU.}
  \label{tab:arch}
  \centering
  \begin{tabular}{lllll}
    \toprule
    X & A & B & C & D\\
    \midrule
    Conv(32,3x3,1) & Conv(64,5x5,1)  & Dropout(0.2)           & Conv(128,3x3,1)  & FC(200) \\
    Conv(32,3x3,1)+MaxPool & Conv(64,5x5,2) & Conv(64,8x8,2)   & Conv(64,3x3,2)   & Dropout(0.5)\\  
    Conv(64,3x3,1)   & Dropout(0.25)         & Conv(128,6x6,2)  & Dropout(0.25)          & FC(200)\\ 
    Conv(64,3x3,1)+MaxPool & FC(128)         & Conv(128,5x5,1)  & FC(128)          & Dropout(0.5)\\  
    FC(200) + Dropout(0.5) & Dropout(0.5)          & Dropout(0.5)           & Dropout(0.5)           & FC(10)+Softmax\\
    FC(200,10)+Softmax   & FC(10)+Softmax        & FC(10)+Softmax         & FC(10)+Softmax         & \\

    \bottomrule
  \end{tabular}

\end{table}

\begin{table}
\small
  \caption{MNIST accuracy results on model transfer and ensemble
    against white-box attacks. X/A means auto-encoder is trained on
    model X, but used on model A. \{XAB\} / C means auto-encoder is
    ensemble trained on three model X,A,B, and evaluated on model C.}
  \label{tab:mnist-transfer}
  \centering
  \begin{tabular}{lrrrrr|rrrr}
    \toprule
    & \multicolumn{5}{c}{Transfer}  & \multicolumn{4}{c}{Ensemble}                   \\
    \cmidrule(r){2-6} \cmidrule(r){7-10}
    
    Attacks & X/X & X/A & X/B & X/C & X/D & \{XAB\}/A & \{XAB\}/B & \{XAB\}/C & \{XAB\}/D \\
    \midrule
    
No attack & 99.1 & 98.1 & 96.8 & 96.5 & 79.7 &  99.0 & 99.1 & 97.9 & 95.0\\
FGSM $\ell_\infty$ & 96.9 & 93.1 & 92.1 & 93.1 & 73.8 &  97.1 & 95.8 & 94.7 & 88.5\\
PGD $\ell_\infty$ & 94.8 & 90.7 & 88.2 & 90.9 & 71.8 &  95.7 & 93.8 & 93.4 & 85.4\\
CW $\ell_2$ & 82.7 & 76.9 & 71.7 & 77.9 & 50.2 &  85.0 & 81.8 & 81.2 & 69.7\\
    \bottomrule
  \end{tabular}
\end{table}

In this section, we evaluate whether auto-encoders trained on one
model can be used to protect other models, i.e. transferability. In
addition to the CNN we used in previous experiment (denoted as model
$X$), we evaluate 4 alternative models that were used
in~\cite{2018-ICLR-Samangouei-Defense}, denoted model A,B,C,D. The
architecture of the all models used is shown in Table
\ref{tab:arch}. There is a ReLU activation layer after each
convolutional layer. All base models are pre-trained with clean
images. For transfer experiment, we train our auto-encoder on model X,
and evaluate on model A,B,C,D. We also evaluate the ensemble
training. In particular, the auto-encoder is trained using
Eq.~\ref{eq:ensemble} on model X,A,B, and tested on model A,B,C,D.


The results are shown in Table \ref{tab:mnist-transfer}.  In the left
half of Table~\ref{tab:mnist-transfer}, we can see the trained
auto-encoders can transfer to protect other models very well. The
accuracy decreases up to 11\% when transferring from model X to model
A,B,C. However, the accuracy drops 20-30\% when transferring from
model X to model D. This is reasonable considering there is a large
architecture change from CNN to FC-only network. In the right half of
Table~\ref{tab:mnist-transfer}, we show the ensemble training across
different CNN models. In all cases, the ensemble boosted the
transferability by a large margin, for example, although not trained
on model C, performance of ``XAB/C'' against all attacks is
considerably better than that of ``X/C''.  Interestingly, the
auto-encoder trained on X,A,B can work very well on model D despite
the dramatic change of model architecture.

\subsection{CIFAR10 results}

We observe the accuracy drop on clean data on AdvAE as well as
It-Adv-train.  This shows the trade-off of robustness and accuracy,
which is also discussed by several recent
papers~\cite{2019-ICLR-Tsipras-Robustness,2018-ECCV-Su-Robustness}.
We didn't compare with DefenseGAN on CIFAR10, as DefenseGAN does not
claim robustness against CIFAR10. We leave the transferability
experiment of models on CIFAR10 as a future work, because CIFAR10
requires much more complicated models such as DenseNet and
ResNet. Transferability between these models is a challenging research
topic.

\begin{table}
\small
  \caption{CIFAR10 accuracy results on three threat models.}
  \label{tab:cifar10}
  \centering
  \begin{tabular}{lr|rr|rr|r}
    \toprule
    & & oblivious & black-box & \multicolumn{3}{c}{white-box}                   \\
    \cmidrule(r){3-3} \cmidrule(r){4-4} \cmidrule(r){5-7}
    Attacks & No defense & AdvAE & AdvAE & \textbf{AdvAE} & HGD & It-Adv-train\\
    \midrule
No attack & 89.2 &  &  & 63.5 & 83.3  & 71.1\\
FGSM $\ell_\infty$ & 22.6 & 63.1 & 60.7 & \textbf{50.8} & 18.1  & \textbf{52.1}\\
PGD $\ell_\infty$ & 10.4 & 62.8 & 60.9 & \textbf{43.6} & 9.7 &  \textbf{48.2}\\
CW $\ell_2$ & 0.0 & 62.7 & 61.3 & 0.2 & 0.0  & 0.0\\
    \bottomrule
  \end{tabular}
\end{table}

The CIFAR10 experiment results are shown in Table
\ref{tab:cifar10}. As a baseline, the undefended model has poor
accuracy on all attacks, while AdvAE performs well against all attacks
in oblivious and black-box attacks.  For white-box attacks, we observe
similar pattern as MNIST results for $\ell_\infty$-bounded attacks,
the FGSM and PGD attacks. In particular, AdvAE achieves 50.8\% and
43.6\% accuracy against FGSM and PGD attack, respectively. This is
on-par with the performance of It-Adv-train performing on the
underlying CNN model, showing again the validity of applying
adversarial training on-demand to only the auto-encoder. Lastly, in
contrary to MNIST dataset, results on CW $\ell_2$-bounded attack
suggest that iterative adversarial training in general (and hence our
AdvAE model) does not offer additional white-box security against the
$\ell_2$-bounded attack on more challenging dataset like CIFAR10.

\section{Related work}

Our work builds upon iterative adversarial training. Adversarial
training~\cite{2015-ICLR-Goodfellow-Explaining} protects a model
against adversarial perturbation by augmenting training data with
adversarial examples. Original adversarial training generates the
adversarial examples once before training. As a result the trained
model is still vulnerable to white-box attacks. Iterative adversarial
training~\cite{2017-ICLR-Kurakin-Adversarial,2018-ICLR-Madry-Towards}
generates adversarial examples during each iteration of training, and
is shown to be successful against white-box attacks. However, such
iterative training scheme introduces much computational overhead. To
make things worse, current adversarial training cannot protect a
pre-trained models without modifying it, thus such computational
overhead needs to be paid whenever a model needs to be hardened.
Ensemble adversarial training~\cite{2018-ICLR-Tramer-Ensemble}
augments the training data with adversarial examples generated on
other pre-trained models.  Like adversarial training, this technique
trains the base classifier, thus cannot be transferred to protect
other models. Feature denoising~\cite{2018-Preprint-Xie-Feature} also
uses adversarial training to enhance white-box robustness. But they
integrate denoise blocks between the convolutional layers in the base
model, and thus is model specific and lack of transferability.




There has been a line of work of purifying-based approach where the
defense mechanism purifies the adversarial examples into benign ones,
without modifying the underlying models being protected. One kind of
such work is by adding denoisers,
e.g. MagNet~\cite{2017-SIGSAC-Meng-Magnet} and
HGD~\cite{2018-CVPR-Liao-Defense}. However, as shown
by~\cite{2017-Preprint-Carlini-Magnet, 2018-ICML-Athalye-Obfuscated},
none of them achieve white-box security.  In particular, HGD by Liao
et al.~\cite{2018-CVPR-Liao-Defense} is very similar to us, in the
sense that they also add auto-encoders that are trained using
adversarial examples. However, same as original adversarial training,
the adversarial examples are pre-generated against the base model.  As
a result, the model is still vulnerable to white-box
attacks~\cite{2018-Preprint-Athalye-Robustness}.
Defense-VAE~\cite{2018-Preprint-Li-Defense} uses VAE instead of
denoising auto-encoders. Same as HGD, Defense-VAE pre-generates
adversarial examples, and does not claim white-box security, but
oblivious one.

Another kind of purifying-based approach builds upon generative
models~\cite{2018-ICLR-Samangouei-Defense, 2017-Preprint-Ilyas-Robust,
  2018-ICLR-Song-Pixeldefend}.  These approaches train a generative
model $G$ for the clean data distribution, and purify an adversarial
example $x$ by finding the closest data point to $x$ on the
generator's manifold $G(z)$. However, such models suffer from
white-box attack in two perspectives. First, as shown by previous
work~\cite{2018-ICML-Athalye-Obfuscated, 2017-Preprint-Ilyas-Robust},
the adversarial examples exist on the generator's manifold, and thus
the underlying principle of these approaches has been shown to provide
0\% accuracy against white-box
attacks~\cite{2018-ICML-Athalye-Obfuscated}. Second, note that the
white-box robustness of these approaches comes from the generator
manifold projection step, where an optimization loop caused the
gradients to vanish. However, this can be simply bypassed by BPDA
algorithm~\cite{2018-ICML-Athalye-Obfuscated} that approximates the
vanished gradient as 1, as it is supposed to be identity mapping. Such
gradient approximation has been shown to reduce the white-box accuracy
significantly~\cite{2018-ICML-Athalye-Obfuscated}.  Ilyas et
al.~\cite{2017-Preprint-Ilyas-Robust} proposed to futher robostify the
generative models through adversarial training on the base classifier.
This is different from our objective, where we do not wish to modify
the underlying classifier.






\section{Conclusion}

In this paper, we propose a novel purifying-based defense approach
that applies iterative adversarial training to auto-encoders to
protect neural models against adversarial examples without modifying
the base models. We empirically show that our proposed method is the
state-of-the-art purifying-based approach against white-box
attacks. Additionally, we show that our auto-encoders can be directly
applied to protect other models very well, and ensemble training
further boost the transferability.



\medskip


\small
\bibliographystyle{plain}
\bibliography{index}



\end{document}